\documentclass[runningheads]{llncs}

\usepackage{eccv}

\usepackage{eccvabbrv}

\usepackage{graphicx}
\usepackage{booktabs}

\usepackage[accsupp]{axessibility}  

\usepackage{hyperref}

\usepackage{orcidlink}

\usepackage{multirow}
\usepackage[table]{xcolor} 
\usepackage{pifont}        
\usepackage{wrapfig}
\newcommand{\xmark}{\ding{55}}%
\begin{document}

\title{OneDrive: Unified Multi-Paradigm Driving with Vision-Language-Action Models}

\titlerunning{OneDrive}

\author{Yiwei Zhang\inst{1,2,3,4} \and Xuesong Chen\inst{4}$^{,\dag}$ \and Jin Gao\inst{1,2}$^{, *}$ \and Hanshi Wang \and Fudong Ge\inst{1,2} \and Weiming Hu\inst{1,2,5} \and Shaoshuai Shi\inst{4}$^{, *}$ \and Zhipeng Zhang\inst{3}$^{, *}$
\\$^*$ Corresponding authors, $^{\dag}$ Project Leader
}

\authorrunning{Y.Zhang et al.}

\institute{
State Key Laboratory of Multimodal Artificial Intelligence Systems, CASIA \and
School of Artificial Intelligence, University of Chinese Academy of Sciences
\and
AutoLab, School of Artificial Intelligence, Shanghai Jiao Tong University\\
\and Voyager Research, Didi Chuxing \and 
School of Information Science and Technology, ShanghaiTech University
\email{\{ywzhang.ai,shuaishaocs\}@gmail.com, jin.gao@nlpr.ia.ac.cn, zhipeng.zhang.cv@outlook.com}
}

\maketitle

\begin{abstract}
Vision-Language Models (VLMs) excel at autoregressive text generation, yet end-to-end autonomous driving requires multi-task learning with structured outputs and heterogeneous decoding behaviors, such as autoregressive language generation, parallel object detection and trajectory regression. To accommodate these differences, existing systems typically introduce separate or cascaded decoders, resulting in architectural fragmentation and limited backbone reuse.
In this work, we present a unified autonomous driving framework built upon a pretrained VLM, where heterogeneous decoding behaviors are reconciled within a single transformer decoder. We demonstrate that pretrained VLM attention exhibits strong transferability beyond pure language modeling.
By organizing visual and structured query tokens within a single causal decoder, structured queries can naturally condition on visual context through the original attention mechanism. 
Textual and structured outputs share a common attention backbone, enabling \textit{stable joint optimization across heterogeneous tasks}.
Trajectory planning is realized within the same causal LLM decoder by introducing structured trajectory queries. This unified formulation enables planning to share the pretrained attention backbone with images and perception tokens.
Extensive experiments on end-to-end autonomous driving benchmarks demonstrate state-of-the-art performance, including 0.28 L2 and 0.18 collision rate on nuScenes open-loop evaluation and competitive results (86.8 PDMS) on NAVSIM closed-loop evaluation. The full model preserves multi-modal generation capability, while an efficient inference mode achieves approximately 40\% lower latency. 
Code and models are available at \url{https://github.com/Z1zyw/OneDrive}

\end{abstract}

\section{Introduction}
\label{sec:intro}

Recent breakthroughs in Vision Language Models (VLMs)~\cite{Qwen2-VL,Qwen2.5-VL,chen2024internvl,zhu2025internvl3} highlight their extraordinary multimodal reasoning capabilities. This success naturally inspires the pursuit of Vision Language Action (VLA) models for autonomous driving~\cite{zhou2025autovla,li2025recogdrive}. Yet integrating these foundational models into driving systems typically demands intricate 3D structural modifications like Bird Eye View (BEV) modeling~\cite{jiang2024senna,han2025percept}. Such severe architectural deviations from native VLMs hinder the possibility of joint training with massive general domain data, fundamentally bottlenecking model scalability. Conversely, attempting to preserve the original architecture exposes a critical dilemma. When structural changes are kept minimal, the model falls significantly behind leading methods on diverse heterogeneous driving tasks. Although previous works manage to mask this deficiency by relying heavily on point cloud inputs, their performance degrades sharply under pure vision settings. This degradation is highly undesirable since pure vision settings naturally align with the original VLM architecture~\cite{han2025percept}. Furthermore, given that these models are inherently constrained to output information in \textbf{textual form}, bridging the massive gap between pure text generation and complex driving outputs remains a formidable challenge. Then,
\begin{quote}
\textit{Can a single pretrained multimodal model simultaneously handle diverse E2E autonomous driving tasks with various forms of output, while maintaining coherent textual generation and generalizing to generative modeling paradigms?}
\end{quote}

\begin{wraptable}{r}{0.5\textwidth}
\vspace{-30pt}
\centering
\small
\caption{Diagnostic study on transferring pretrained LLM weights to a parallel decoder. \xmark~means random initialize.}
\begin{tabular}{l|cccc}
\toprule
VLM   & ~Attn~ & FFN~ & NDS$\uparrow$ \\
\midrule
\multirow{4}{*}{InternVL3-1B~\cite{zhu2025internvl3}} &  \checkmark & \checkmark & 31.95 \\
  & \checkmark & \xmark & \textbf{32.05} \\
    & \xmark & \checkmark & 29.90 \\
  & \xmark & \xmark & 31.48 \\
\midrule
\multirow{4}{*}{Qwen2.5-VL-3B~\cite{Qwen2.5-VL}}  &  \checkmark & \checkmark & 27.14 \\
     &  \checkmark & \xmark & \textbf{31.37} \\
     &  \xmark & \checkmark & 27.95 \\
     &  \xmark & \xmark & 30.15 \\
\bottomrule
\end{tabular}
\label{tab:weight_transfer}
\vspace{-15pt}
\end{wraptable}



Tackling this question requires confronting the inherent complexity of autonomous driving, where interdependent tasks demand distinct input and output formats. Perception and planning typically employ parallel decoders with specific attention mechanisms for simultaneous predictions~\cite{detr,zhu2020deformable,zhang2026integrating,bevformer,bevformerv2,MapTR,maptrv2,zhang2024vq,vad,chen2024vadv2,PARAdrive,chitta2022transfuser}. Conversely, textual reasoning relies on sequential autoregressive decoding~\cite{sima2023drivelm,chen2025asynchronous,wang2024omnidrive,jiang2024senna} (see Fig.~\ref{fig:arch-paral-ar} (b)). This profound structural gap forces traditional approaches to use isolated~\cite{PARAdrive} or cascaded~\cite{uniad,vad,jia2025drivetransformer,wang2024omnidrive,chen2025asynchronous,drivevlm} decoders. Such fragmented designs strictly prevent the unified application of pretrained weights and restrict information flow across tasks. Consequently, engineering a singular architecture that harmonizes these diverse modalities while preserving the native VLM structure remains a formidable open challenge.

To determine whether native pretrained VLM models can actually overcome this challenge and support diverse driving tasks, we conduct a preliminary diagnostic study. We adapt pretrained VLM weights into a parallel decoder using the first six layers of the language model, inspired by DETR~\cite{detr} and StreamPETR~\cite{streampetr}. Specifically, we freeze the visual backbone and map each pretrained self attention module to the corresponding cross attention in the parallel decoder. For the feedforward networks, we compare \textit{pretrained initialization} against \textit{random initialization}, alongside a fully random baseline for all modules to ensure completeness. As Tab.~\ref{tab:weight_transfer} demonstrates, attention weights transfer effectively across tasks. Conversely, reusing feedforward weights provides limited benefits and occasionally causes severe performance degradation alongside training instability (31.37 $\to$ 27.14). These observations reveal that feedforward networks optimized strictly for text generation lack the flexibility required for heterogeneous downstream tasks. However, attention modules pretrained to align visual and textual tokens retain remarkable transferability. This success likely stems from their ability to capture fundamental correspondence patterns between queries and visual features. Inspired by this crucial insight, we investigate whether such transferable attention mechanisms can model a broader range of relationships, ultimately enabling the construction of a unified multitask decoder.

\begin{wrapfigure}{r}{0.5\textwidth}
    \centering
    \small
    \vspace{-20pt}
    \includegraphics[width=0.95\linewidth]{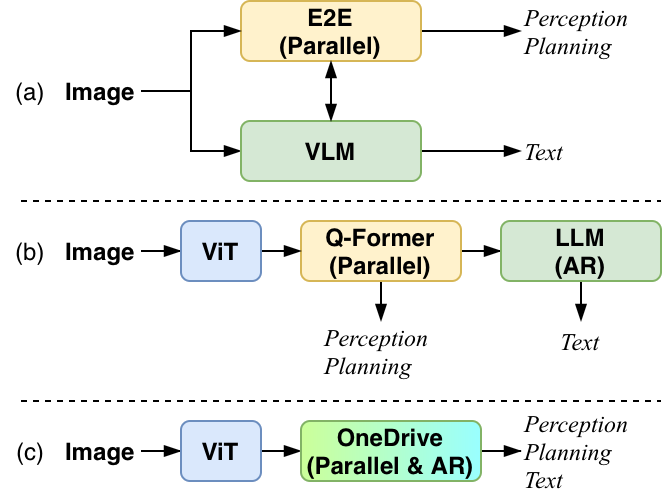}
    \caption{(a) dual-system design with separate decoders; (b) Q-Former–style cascaded decoding; (c) our unified single-decoder framework handling both within one transformer.}
    \label{fig:arch-vlm-cmp}
    \vspace{-15pt}
\end{wrapfigure}



\textbf{OneDrive}~(see Fig.~\ref{fig:arch-ours}) is our definitive answer.
It is a unified multitask framework elegantly leverages pretrained attention to support heterogeneous decoding behaviors within a single transformer decoder. Specifically, we organize visual tokens and structured query tokens into a unified sequence. This arrangement allows query tokens to deeply condition on visual context through the pretrained causal attention of the language model. To enable structured prediction, we augment the shallow layers with dedicated attention among perception query tokens. This modification enhances comprehensive perception capabilities while keeping the foundational backbone attention strictly unchanged. Furthermore, we insert task-specific feedforward networks to execute necessary feature transformations. This elegant design allows autoregressive textual generation alongside parallel perception and planning to share a common attention backbone. Consequently, our design enables stable joint optimization across diverse tasks without introducing isolated transformer branches. Moreover, the unified sequence and shared attention structure inherently facilitate seamless interaction across all driving tasks.

Extensive experiments on end-to-end autonomous driving benchmarks demonstrate that our framework achieves state-of-the-art performance, including 0.28 $L_2$ error and 0.18 collision rate on nuScenes open-loop evaluation and competitive results on NAVSIM closed-loop evaluation. The full model preserves multi-modal generation capability, while a truncated inference mode that forwards only the early layers reduces latency by approximately 40\% (264→156 ms) for planning-focused deployment.

The main contributions of this work are as follows:
\begin{enumerate}
    \item We show that pretrained VLM causal attention, originally trained to capture text–image or text–text relations, can be adapted through training to model relations between queries and visual, whereas FFNs pretrained for text generation are difficult to transfer.
    \item We propose OneDrive, a unified archicture for multitask, enabling a single transformer decoder to perform both autoregressive text generation and parallel perception and planning tasks.
    \item Extensive experiments on nuScenes and NAVSIM demonstrate the effectiveness of our approach, achieving state-of-the-art performance.
\end{enumerate}

\section{Related Work}
\label{sec:rel}

\subsubsection{Unified Architectures for End-to-end Autonomous Driving.}

\begin{wrapfigure}{r}{0.5\textwidth}
    \centering
    \small
    \vspace{-20pt}
    \includegraphics[width=0.95\linewidth]{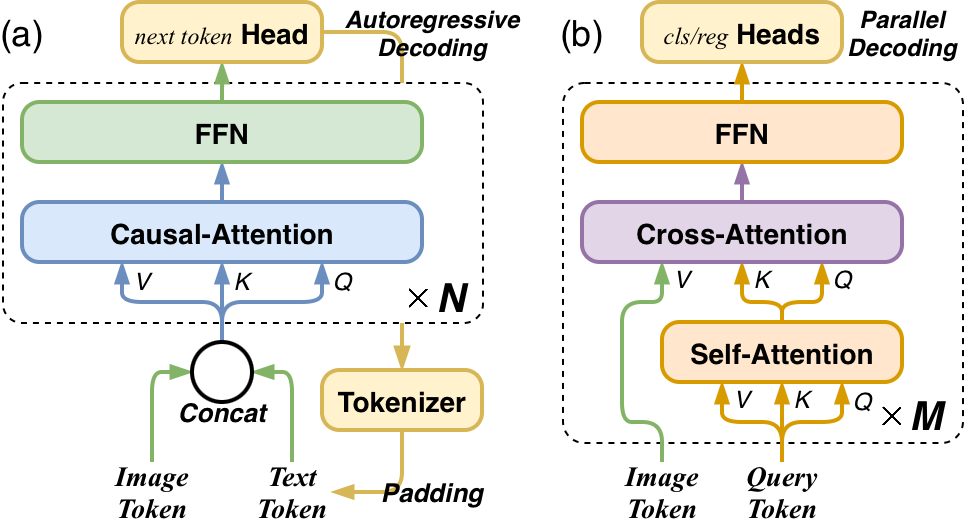}
    \caption{Two representative decoding paradigms: (a) an autoregressive decoder, (b) a parallel decoder. Existing end-to-end multi-task autonomous driving models typically organize heterogeneous decoders either in a cascaded manner or in parallel.}
    \label{fig:arch-paral-ar}
    \vspace{-15pt}
\end{wrapfigure}

Autonomous driving requires the integration of multiple interdependent tasks, including perception, prediction, and planning, such as 3D object detection~\cite{huang2021bevdet,bevformer,bevformerv2,bevfusion}, lane detection~\cite{MapTR,maptrv2,maptracker}, BEV segmentation~\cite{bevfusion,zhang2024vq}, occupancy prediction~\cite{zheng2024monoocc,occworld}, and trajectory planning. Early systems decomposed these tasks into separate modules with predefined interfaces, which limited information flow and hindered global optimization.
Recent unified frameworks~\cite{uniad,vad,chen2024vadv2,PARAdrive,jia2025drivetransformer}, often referred to as conventional end-to-end driving models~\cite{GE2EAD}, advocate for the joint optimization of perception, mapping, motion forecasting, and planning within a shared backbone. These approaches demonstrate that multi-task learning effectively enhances cross-task consistency and overall computational efficiency. However, these systems do not incorporate language modeling into the unified architecture. Notably, the autoregressive decoding paradigm of language models (Fig.~\ref{fig:arch-paral-ar}(b)) differs fundamentally from the parallel structured decoders (Fig.~\ref{fig:arch-paral-ar}(a)) commonly adopted in end-to-end driving frameworks.



\subsubsection{Vision-Language Models in Autonomous Driving.}


Recent studies have explored integrating vision-language models into autonomous driving to enhance scene understanding, instruction following, and interpretability. Dual-system designs~(see Fig.~\ref{fig:arch-vlm-cmp}(a)), such as Senna~\cite{jiang2024senna} and DriveVLM~\cite{drivevlm}, employ separate decoders for structured prediction and language generation. Other works, including OmniDrive~\cite{wang2024omnidrive}, Orion~\cite{fu2025orion} and SOLVE~\cite{chen2025asynchronous}, adopt cascaded architectures that combine structured decoders with autoregressive LLM decoders~(see Fig.~\ref{fig:arch-vlm-cmp}(b)). In contrast, some approaches~\cite{hwang2024emma, xing2024openemma} directly formulate all tasks as text generation using a single autoregressive decoder.
Despite these efforts, existing methods typically organize heterogeneous decoders either in parallel or in cascaded forms. Conventional end-to-end driving models rely primarily on parallel decoding, while large language models operate in an autoregressive manner. The structural discrepancy between autoregressive token generation and parallel structured prediction remains largely unresolved.

In contrast to prior VLM-based driving works that treat language as an auxiliary modality and unified driving frameworks that integrate tasks through shared representations, our work focuses on unifying heterogeneous driving outputs at the decoding level. We adopt a token-centric formulation based on a pretrained VLM, enabling a single Transformer decoder to generate structured perception, planning trajectories, and textual responses within a unified causal attention framework.



\begin{figure*}[t]
\centering
\includegraphics[width=0.98\linewidth]{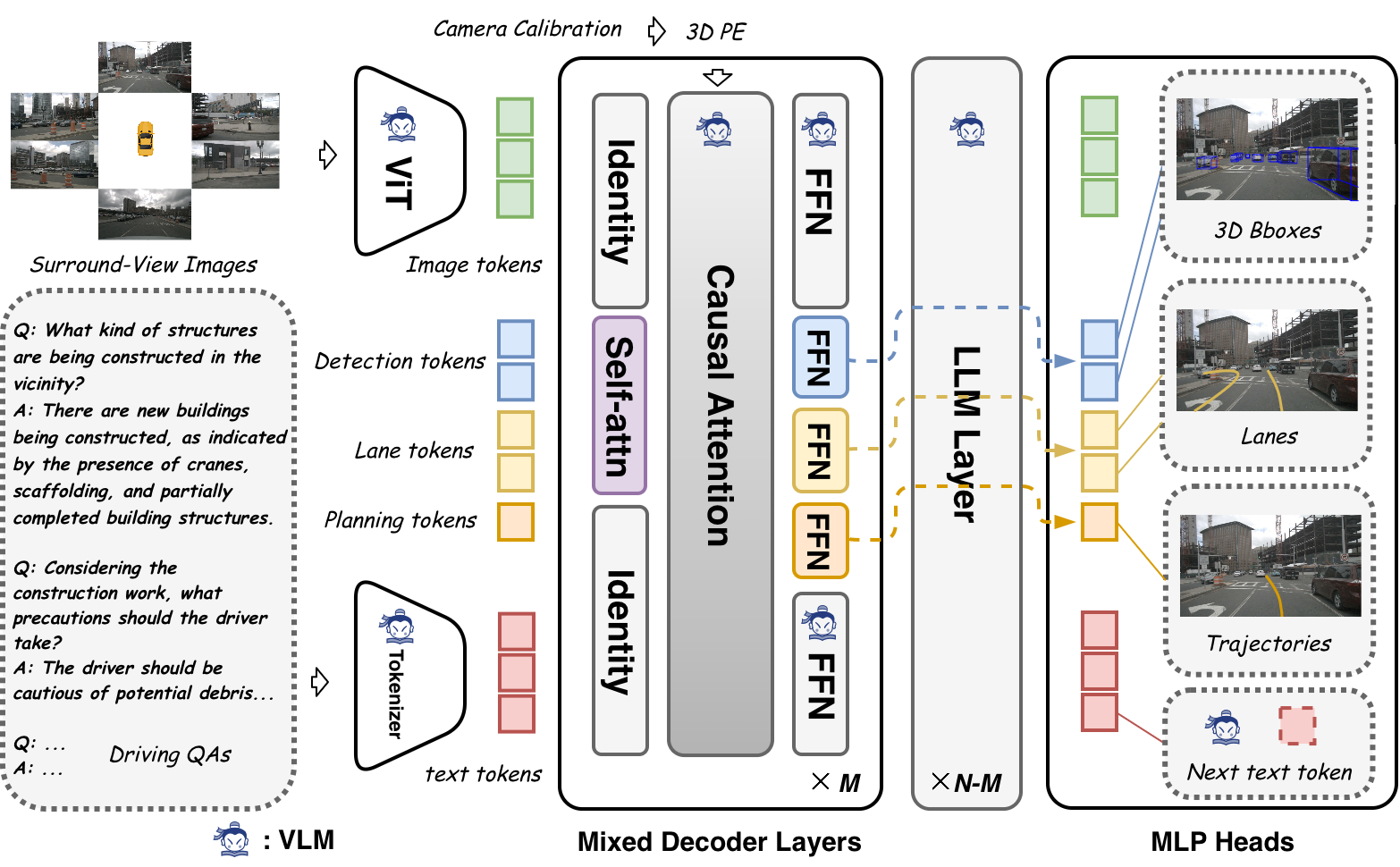} 
\caption{Architecture of OneDrive. Surround-view images are encoded into image tokens by a ViT and concatenated with structured query tokens for detection, lane estimation, and planning, as well as text tokens. The unified token sequence is processed by mixed decoder layers built upon the pretrained LLM causal attention. Perception query tokens are augmented with additional self-attention and task-specific feed-forward networks, while the backbone attention remains unchanged. Task-specific MLP heads decode 3D bounding boxes, lanes, and trajectories in parallel, alongside autoregressive next-token prediction for textual generation, enabling unified multi-task learning within a single transformer decoder.}
\label{fig:arch-ours}
\end{figure*}
\section{OneDrive}
\label{sec:method}

Experiments in Table~\ref{tab:weight_transfer} show that adapting pretrained VLMs to heterogeneous driving structures and tasks is nontrivial: attention modules transfer well, but feedforward layers often degrade performance. These findings highlight the need for careful architectural and training design for a unified multitask decoder.

Fig.~\ref{fig:arch-ours} illustrates the overall architecture of OneDrive. We build upon a pretrained vision-language model and unify perception, planning, and language generation within a single Transformer decoder. Visual tokens, structured query tokens, and text tokens are concatenated into a shared sequence and processed by the pretrained causal attention. To accommodate heterogeneous prediction paradigms, we introduce task-specific adaptations, including additional self attention among structured perceptual queries and task-specific feed-forward networks, while preserving the original attention backbone.

\subsection{Problem Formulation}

We formulate end-to-end autonomous driving as a multi-task prediction problem over heterogeneous output spaces. Given multi-view images $\mathcal{I}$ and optional textual inputs, the system is required to jointly predict \textbf{(i)} structured perception outputs, such as 3D bounding boxes and lane structures; \textbf{(ii)} continuous trajectory predictions for planning; and \textbf{(iii)} optional textual descriptions. These tasks differ not only in output representation but also in decoding paradigms, including parallel query-based prediction and autoregressive token generation.

\subsection{Unified Token Representation}

Instead of introducing task-specific decoders, we organize heterogeneous outputs using a unified token-based formulation.
Structured prediction tasks, such as 3D object detection and lane estimation, are represented by fixed sets of learnable query tokens. Planning is modeled with dedicated planning tokens, while textual outputs follow standard autoregressive token generation.
All tokens are concatenated into a shared sequence processed by a single Transformer decoder:

\[
\mathbf{Z} = [\mathbf{X}_{img}, \mathbf{Q}_{det}, \mathbf{Q}_{lane}, \mathbf{Q}_{plan}, \mathbf{X}_{text}],
\]
where $\mathbf{X}_{img}$ denotes image tokens, $\mathbf{Q}_{\cdot}$ denote task-specific query tokens, and $\mathbf{X}_{text}$ represents text tokens. This unified token organization enables heterogeneous prediction paradigms to be handled within a shared decoder architecture. 

Similar to SOLVE~\cite{chen2025asynchronous} and OmniDrive~\cite{wang2024omnidrive}, we adopt StreamPETR~\cite{streampetr} to organize the detection and lane queries. Specifically, $\mathbf{Q}_{det}$ and $\mathbf{Q}_{lane}$ are constructed following the query formulation of StreamPETR. For planning, we allocate one planning query per future timestep. Each planning query is initialized with an anchor trajectory derived from VAD~\cite{vad}. In addition, an ego-status embedding is prepended to the planning queries, forming the complete $\mathbf{Q}_{plan}$.

\subsection{Mixed Decoder Layers}

Although all tokens share a unified representation space, heterogeneous tasks entail distinct interaction patterns and feature transformation requirements. To preserve the transferable relational modeling capacity of pretrained VLMs while enabling parallel structured prediction, mixed decoder layers are introduced in the shallow stages, retaining the original causal attention while incorporating minimal task-specific adaptations. The query outputs for structured prediction are directly taken from these shallow mixed layers, whereas the deeper layers remain unchanged from the pretrained decoder.

\subsubsection{Causal Attention as Conditional Interface}

At the core of each decoder layer, we retain the pretrained LLM causal self-attention. All tokens, including image tokens, perception queries, planning queries, and text tokens, are processed under a shared causal mask.

To enhance spatial modeling for perception and planning, we follow StreamPETR and incorporate 3D positional embeddings~\cite{liu2022petr} for image tokens and structured query tokens, applied after RoPE~\cite{su2024roformer}, while text tokens remain unchanged.
Specifically, for image tokens $\mathbf{X}_{img}$ and structured query tokens $\mathbf{Q}_{task}$, the attention projections are computed as
$$
\mathbf{Q} = \text{RoPE}(\mathbf{X}W_q) + e^{3D}, \quad
\mathbf{K} = \text{RoPE}(\mathbf{X}W_k) + e^{3D},
$$
where $\mathbf{X} \in \{\mathbf{X}_{img}, \mathbf{Q}_{det}, \mathbf{Q}_{lane}, \mathbf{Q}_{plan}\}$ and $e^{3D}$ denotes the corresponding 3D positional embedding. Text tokens $\mathbf{X}_{text}$ follow the original RoPE formulation without additional 3D positional encoding. For the query tokens, we remove the residual connection in the causal attention module, which stabilizes training when adapting from pretrained VLM weights.

This shared attention backbone serves as a unified conditional modeling interface across modalities and tasks. Since query tokens are placed after image tokens, they naturally condition on visual context via causal attention without requiring explicit cross-attention modules. Planning tokens are further appended after perception queries, enabling implicit conditioning on perception features through the same attention mechanism.


\subsubsection{Query Interaction and Task-Specific Transformation}

Autoregressive causal attention enforces sequential dependencies, which is not ideal for parallel structured prediction. To support end-to-end perception~\cite{detr,sun2021makes}, we introduce an additional self-attention block applied only among structured query tokens:
$$
{\mathbf{Q_{perception}}} =
\text{SelfAttn}_{q}(\mathbf{Q_{perception}})=\text{SelfAttn}_{q}([\mathbf{Q_{det}},\mathbf{Q_{lane}}]),
$$
This module enables bidirectional interaction within each query group, facilitating parallel reasoning while leaving the backbone causal attention unchanged. Furthermore, we replace the pretrained language-modeling feed-forward networks (FFNs) for structured queries with task-specific FFNs:

$$\mathbf{Q}' = \text{FFN}_{t}(\tilde{\mathbf{Q}}),$$
where $t \in \{\text{det}, \text{lane}, \text{plan}\}$. Text tokens continue to use the original pretrained FFNs. This design preserves transferable attention while allowing task-dependent feature transformations.

\subsection{Multi-Stage Training}
\label{subsec:multi-stage}

We adopt a three-stage training strategy for stable adaptation from pretrained VLMs to multi-task driving. In the first stage (\textbf{Perception–language pretraining}), we freeze the ViT encoder and train the mixed decoder with perception and text-generation objectives. The causal attention is fully fine-tuned, the LLM decoder is adapted via LoRA, and the perception-specific self-attention, perception FFNs, and MLP heads are randomly initialized and optimized. Gradients are driven by perception and language modeling losses, denoted as $\mathcal{L}_{\text{pretrain}} = \lambda_{perc} \mathcal{L}_{\text{perc}} + \mathcal{L}_{\text{text}}$. In the second stage (\textbf{Planning adaptation}), planning tokens are introduced. We optimize planning queries, the planning FFN, and the planning MLP head in an end-to-end manner, while continuing LoRA adaptation of the LLM decoder. Perception-specific modules remain fixed. Gradients are provided by planning and language modeling loss  $\mathcal{L}_{\text{adaptation}} = \lambda_{plan} \mathcal{L}_{\text{plan}} + \mathcal{L}_{\text{text}}$. In the final stage (\textbf{Joint finetuning}), all modules, including the ViT encoder, are jointly fine-tuned under the combined perception, planning, and text objectives, with the overall training loss $\mathcal{L}_{\text{joint}} = \lambda_{perc} \mathcal{L}_{\text{perc}} + \lambda_{plan} \mathcal{L}_{\text{plan}} + \mathcal{L}_{\text{text}}$, enabling full end-to-end optimization of the framework.
For object detection~\cite{zhu2020deformable,zhang2026integrating,streampetr,bevformer}, deeply supervision~\cite{lee2015deeply} is often applied to propagate gradients at every layer, where each decoder layer computes its own loss. Similarly, we employ deeply supervision for the planning task.

\section{Experiments}
\label{sec:exp}

\subsection{Implement Details}
\subsubsection{Datasets and Evaluation.}
We conduct experiments on two autonomous driving benchmarks: \textbf{nuScenes} and \textbf{NAVSIM}.

\textbf{\underline{nuScenes}}~\cite{nuscenes} contains 1,000 urban driving scenes, each approximately 20 seconds long, captured with six surround-view cameras and a LiDAR sensor. The dataset provides 3D bounding box annotations for 10 object categories, along with high-definition semantic maps. We follow the standard split with 700 training, 150 validation, and 150 test scenes. For perception supervision, we use 3D detection annotations and lane annotations derived from OpenLaneV2~\cite{wang2023openlane}. To enrich language and reasoning signals, we additionally leverage the OmniDrive extension~\cite{wang2024omnidrive}, which augments nuScenes with QA-style annotations spanning perception, prediction, and planning.
For planning in the open-loop setting, we measure trajectory accuracy using the L2 displacement error and assess safety using collision rate. 


\textbf{\underline{NAVSIM}}\cite{dauner2025navsim} is a planning-oriented benchmark built upon OpenScene~\cite{openscene2023}, a redistribution of nuPlan\cite{caesar2021nuplan}. The dataset is divided into 1,192 training scenes (navtrain) and 136 evaluation scenes (navtest). Compared with nuScenes, NAVSIM focuses more heavily on interactive and safety-critical planning scenarios. Metrics include average displacement error, collision rate, and the official NAVSIM score PDMS, which jointly reflect safety, rule compliance, and driving efficiency.

\subsubsection{Training.}
On the nuScenes benchmark, our model is built upon InternVL3-1B~\cite{zhu2025internvl3}. We follow the three-stage training strategy described in Sec.~\ref{subsec:multi-stage}, where each stage is trained for 20 epochs with an initial learning rate of $1\times10^{-4}$. Experiments are conducted with a batch size of 64 on $64\times$ NVIDIA H20 GPUs.

On NAVSIM, we initialize InternVL3-2B~\cite{zhu2025internvl3} from the ReCogDrive~\cite{li2025recogdrive} checkpoint, which has been fine-tuned for autonomous driving scenarios. In this setting, we use only planning queries without introducing additional perception queries, while keeping the same configuration as in Stage1. The model is trained with a learning rate of $1\times10^{-4}$ and a batch size of 128 .




\subsection{Main Results}

\subsubsection{Open-loop evaluation in nuScenes.}

\begin{table*}[ht!]
\small
\centering
\vspace{-10pt}
\caption{Performance comparison of different methods on the nuScenes dataset for open-loop planning. Methods are categorized into text-based driving models (top) and action-based driving models (bottom). $^{\dag}$ indicates that the model does not use ego-vehicle status as input.}
{\begin{tabular}[b]{l|c|cccc|cccc}
\toprule[1.5pt]
\multirow{2}{*}{\textbf{Method}} &
\multirow{2}{*}{\textbf{Reference}} &
\multicolumn{4}{c|}{\textbf{L2 ($m$)} $\downarrow$} &
\multicolumn{4}{c}{\textbf{Col. Rate (\%)} $\downarrow$ }\\
& & 1$s$ & 2$s$ & 3$s$ & Avg. & 1$s$ & 2$s$ & 3$s$ & Avg. \\
\midrule

\multicolumn{10}{c}{\textbf{Text-Based Driving Models}} \\
\midrule
DriveVLM~\cite{tian2024drivevlm} & CoRL 2024 & 0.18 & 0.34 & 0.68 & 0.40 & -- & -- & -- & -- \\
DriveVLM-Dual~\cite{tian2024drivevlm} & CoRL 2024  & 0.15 & 0.29 & 0.48 & 0.31 & -- & -- & -- & -- \\
OmniDrive~\cite{wang2024omnidrive} & CVPR 2025 & 0.14 & 0.29 & 0.55 & 0.33 & \textbf{0.00} & {0.13} & 0.78 & 0.30 \\
EMMA~\cite{hwang2024emma} & TMLR & 0.14 & 0.29 & 0.54 & 0.32 & -- & -- & -- & -- \\
EMMA+~\cite{hwang2024emma} & TMLR & \textbf{0.13} & 0.27 & 0.48 & 0.29 & -- & -- & -- & -- \\
ImpromptuVLA~\cite{chi2025impromptu} & NeurIPS 2025  &\textbf{0.13}          & 0.27      & 0.53       &  0.30      & -- & -- & --  &  -- \\
SOLVE-VLM~\cite{Chen_2025_CVPR} & CVPR 2025 & \textbf{0.13} & \textbf{0.25} & \textbf{0.47} & \textbf{0.28} & \textbf{0.00} & 0.16 & \textbf{0.43} & \textbf{0.20} \\
VGGDrive~\cite{wang2026vggdrive} & CVPR 2026 & 0.14 & 0.28 & 0.51 & 0.31 & 0.02 & \textbf{0.10} & 0.55 & 0.22 \\
\midrule

\multicolumn{10}{c}{\textbf{Action-Based Driving Models}} \\
\midrule
UniAD$^{\dag}$~\cite{hu2023planning} & CVPR 2023 & 0.59 & 1.01 & 1.48 & 1.03 & 0.16 & 0.51 & 1.64 & 0.77 \\
VAD-Base$^{\dag}$~\cite{jiang2023vad} & ICCV 2023 & 0.69 & 1.22 & 1.83 & 1.25 & 0.06 & 0.68 & 2.52 & 1.09 \\
BEV-Planner$^{\dag}$~\cite{li2024ego} & CVPR 2024 &  0.30 & 0.52 & 0.83 & 0.55 & 0.10 & 0.37 & 1.30 & 0.59 \\
UniAD~\cite{hu2023planning} & CVPR 2023 & 0.20 & 0.42 & 0.75 & 0.46 & 0.02 & 0.25 & 0.84 & 0.37 \\
VAD-Base~\cite{jiang2023vad} & ICCV 2023 & 0.17 & 0.34 & 0.60 & 0.37 & 0.04 & 0.27 & 0.67 & 0.33 \\
AD-MLP~\cite{admlp} & ArXiv 2023 & 0.15 & 0.32 & 0.59 & 0.35 & \textbf{0.00} & 0.27 & 0.85 & 0.37 \\
BEV-Planner~\cite{li2024ego} & CVPR 2024 & 0.16 & 0.32 & 0.57 & 0.35 & \textbf{0.00} & 0.29 & 0.73 & 0.34 \\
SOLVE-E2E~\cite{Chen_2025_CVPR} & CVPR 2025  & 0.14 & 0.28 & 0.50 & 0.31 & 0.04 & {0.17} & 0.68 & 0.30 \\
ColaVLA~\cite{peng2025colavla} & CVPR 2026 & {0.14} & {0.27} & {0.50} & {0.30} & 0.04 & {0.17} & {0.47} & {0.23} \\
 \rowcolor{gray!20}
OneDrive & --  & \textbf{0.13} & \textbf{0.25} & \textbf{0.46} & \textbf{0.28} & \textbf{0.00} & \textbf{0.12} & \textbf{0.43} & \textbf{0.18} \\
\bottomrule[1.5pt]
\end{tabular}}

\label{tab:sota-nusc}
\end{table*}

As shown in Table~\ref{tab:sota-nusc}, among action-based planners, OneDrive achieves the best overall performance in terms of both accuracy and safety, with the lowest average L2 error (0.28 m) and the lowest average collision rate (0.18\%). Compared with the strongest prior action-based baselines, SOLVE-E2E (0.31m L2; 0.30\% collision) and ColaVLA (0.30m L2; 0.23\% collision), our method reduces the L2 error to 0.28 m and lowers the collision rate by 23\% relative to ColaVLA, indicating more precise and safer trajectory prediction.

Although ColaVLA also avoids autoregressive decoding, it requires forwarding the full LLM and relies on a customized attention mask, which prevents the use of optimized implementations such as FlashAttention~\cite{dao2022flashattention}. In contrast, OneDrive only forwards the shadow LLM layers without introducing custom attention masking, enabling a more lightweight and hardware-efficient integration(see Table~\ref{tab:latency}). Moreover, OneDrive remains competitive with recent text-based VLM planners, while avoiding SOLVE-VLM autoregressive text generation with complex chain of thought.


\subsubsection{Closed-loop evaluation in NAVSIM.}

\begin{table*}[ht!]
\small
\centering
\vspace{-10pt}
\caption{Closed-loop performance on NAVSIM navtest under supervised fine-tuning. Our unified modeling improves planning stability and interaction awareness over prior end-to-end and VLA methods, while reducing inference cost by approximately 40\% through unified causal attention without forwarding the full LLM. $^{\dag}$ means using Lidar as extra input.}
{\begin{tabular}[b]{l|c|cc|ccc|c}
\toprule[1.5pt]
\textbf{Method} & \textbf{Reference} & \textbf{NC$\uparrow$} & \textbf{DAC$\uparrow$} & \textbf{TTC$\uparrow$} & \textbf{Comf.$\uparrow$} & \textbf{EP$\uparrow$} & \textbf{PDMS$\uparrow$} \\
\midrule
Ego Status MLP & - & 93.0 & 77.3 & 83.6 & 100 & 62.8 & 65.6 \\
VADv2~\cite{chen2024vadv2} & ArXiv 2024 & 97.2 & 89.1 & 91.6 & 100 & 76.0 & 65.6 \\
DrivingGPT~\cite{chen2024drivinggpt} &ICCV 2025 & 98.9 & 90.7 & 94.9 & 95.6 & 79.7 & 82.4 \\
UniAD~\cite{uniad} & CVPR 2023 & 97.8 &91.9 &92.9 &100 &78.8 &83.4 \\ 
TransFuser$^{\dag}$~\cite{chitta2022transfuser} & PAMI 2022 & 97.7 &92.8& 92.8 &100 &79.2 &84.0 \\ 
PARA-Drive~\cite{weng2024paradrive} & CVPR 2024 & 97.9 &92.4 &93.0 &99.8 &79.3 &84.0 \\
DRAMA$^{\dag}$~\cite{yuan2024drama} & ArXiv 2024 & 98.0 &93.1 &94.8 &100 &80.1 &85.5 \\ 
Hydra-MDP$^{\dag}$~\cite{hydraMDP} & ArXiv 2024 & 98.3 &96.0 &94.6 &100 &78.7 &86.5 \\
DiffusionDrive$^{\dag}$~\cite{liao2024diffusiondrive} & CVPR 2025 & 98.2 &96.2 &94.7 &100 &82.2 &88.1 \\ 
\midrule
Qwen2.5-VL-8B~\cite{Qwen2.5-VL} & - & 97.8 & 92.1 & 92.8 & 100 & 78.3 & 83.3 \\
InternVL3-8B~\cite{zhu2025internvl3} & - & 97.0 & 92.1 & 91.8 & 100 & 78.9 & 83.3 \\
AutoVLA(SFT)~\cite{zhou2025autovla} & NeurIPS 2025 & 96.9 & 94.4 & 88.1 & 99.9 & 75.8 & 80.5 \\
ReCogDrive(SFT)~\cite{li2025recogdrive} & ICLR 2026 & 98.1 & 94.7 & 94.2  & 100 & 80.9 & 86.5 \\
\rowcolor{gray!20}
Query Decoder Baseliine & -- & -- & -- & -- & -- & -- & 85.0 \\
\rowcolor{gray!20}
OneDrive & -- & 98.4 & 95.2 & 94.9 & 100 & 81.1 & 86.8 \\
\bottomrule[1.5pt]
\end{tabular}}

\label{tab:sota-plan}
\end{table*}
Table~\ref{tab:sota-plan} shows the performance comparison on the NAVSIM navtest benchmark, evaluated using closed-loop metrics, including both state-of-the-art end-to-end driving approaches and existing VLA models under supervised fine-tuning. Rather than solely emphasizing absolute leaderboard ranking, this evaluation focuses on how our unified modeling strategy enhances the closed-loop trajectory planning capability of the base VLM. 
To provide a fair comparison, we additionally include a Query Decoder baseline, which adopts the query-based planning decoder used in ReCogDrive~\cite{li2025recogdrive}. As shown in the table, while this baseline already achieves competitive performance (PDMS 85.0), our method further improves the result to 86.8 PDMS, indicating that the proposed unified causal decoding design is more effective than conventional query-based planning heads.
Notably, our method achieves these gains while reducing the inference cost by approximately 40\% (see Table~\ref{tab:latency}), as it avoids repeatedly forwarding the full LLM. This highlights that the proposed design improves not only closed-loop performance but also computational efficiency, making it more suitable for real-time autonomous driving scenarios.

\subsection{Ablation Study}
\subsubsection{Text Evaluation.}
\begin{wraptable}{r}{0.5\textwidth}
\vspace{-35pt}
\centering
\small
\caption{Comparison with OmniDrive under the same training data.}
\begin{tabular}{l|cccc}
\toprule
\multirow{2}{*}{Method} & \multicolumn{4}{c}{\textbf{L2 ($m$)} $\downarrow$} \\
 & 1$s$ & 2$s$ & 3$s$ & Avg. \\
\midrule
OmniDrive-7B~\cite{li2025recogdrive} & 0.14 & 0.29 & 0.55 & 0.33 \\
\rowcolor{gray!20}
Ours-Text-1B & 0.15 & 0.29 & 0.51 & 0.32 \\
\bottomrule
\end{tabular}

\label{tab:text-cmp}
\vspace{-25pt}
\end{wraptable}
To verify that the proposed framework preserves the language generation capability of the underlying VLM, we compare it with OmniDrive under the same training data setting. Specifically, we train our model using the same dataset as OmniDrive and evaluate the text-conditioned trajectory prediction accuracy. As shown in Table~\ref{tab:text-cmp}, our method achieves comparable or better L2 errors across the prediction horizons, yielding a lower average error (0.32 vs. 0.33). These results indicate that integrating the planning decoder and perception modules does not degrade the language capability of the model, while maintaining competitive performance in text-conditioned driving tasks.

\subsubsection{Disentangling the Role of Text Supervision.}
\begin{table*}[ht!]
\vspace{-30pt}
\small
\centering
\caption{Impact of Text Supervision on Perception and Planning.}
{\begin{tabular}[b]{l|c|c|cccc|cccc}
\toprule[1.5pt]
\multirow{2}{*}{\textbf{Task}} & \multirow{2}{*}{\textbf{Text}} &
\multirow{2}{*}{\textbf{NDS/mAP}} &
\multicolumn{4}{c|}{\textbf{L2 ($m$)} $\downarrow$} &
\multicolumn{4}{c}{\textbf{Col. Rate (\%)} $\downarrow$ }\\
& & & 1$s$ & 2$s$ & 3$s$ & Avg. & 1$s$ & 2$s$ & 3$s$ & Avg. \\
\midrule
3D Det  &  & 32.31/22.47 & -- & -- & -- & -- & -- & -- & -- & -- \\
3D Det  & \checkmark & \textbf{33.94}/\textbf{24.39} & -- & -- & -- & -- & -- & -- & -- & -- \\
E2E &  & -- & \textbf{0.13} & \textbf{0.27} & \textbf{0.51} & \textbf{0.31} & 0.04 & \textbf{0.19} & 0.98 & 0.40 \\
E2E & \checkmark & -- &\textbf{0.13} & \textbf{0.27} & \textbf{0.51} & \textbf{0.31} & \textbf{0.02} & 0.20 & \textbf{0.85} & \textbf{0.36} \\

\bottomrule[1.5pt]
\end{tabular}}
\vspace{-15pt}


\label{tab:abl-wotext}
\end{table*}

As shown in Table~\ref{tab:abl-wotext}, we analyze the effect of retaining text supervision when using a unified causal attention to jointly model perception (3D object detection) or planning. Enabling the text loss leads to small but consistent improvements across tasks. For 3D detection, the NDS/mAP increases slightly after introducing text supervision. For end-to-end planning, the L2 displacement error remains largely unchanged, while the collision rate is reduced (from 0.40 to 0.36), indicating improved driving safety. We attribute this behavior to the optimization characteristics of the pretrained causal attention. The causal attention module is originally pretrained under autoregressive text generation objectives, where its parameter space is shaped to capture sequential semantic dependencies. When adapting it to heterogeneous structured prediction tasks such as object detection and trajectory regression, the optimization process may gradually drift from the regime favored during pretraining. Maintaining a text loss during training helps preserve partial alignment with the original objective, which can provide a mild regularization effect on the shared attention module. This regularization stabilizes the unified causal attention to some extent, benefiting both perception and planning.


\subsubsection{Key Hyperparameter $\lambda_{plan}$.}
\begin{table*}[ht!]
\vspace{-30pt}
\small
\centering
\caption{Ablation on hyperparameter $\lambda_{plan}$.We vary the weight of the planning loss in the third stage (joint training) to study its effect on planning performance.}
{\begin{tabular}[b]{l|c|cccc|cccc}
\toprule[1.5pt]
\multirow{2}{*}{$\lambda_{plan}$} & \multirow{2}{*}{\textbf{NDS/mAP}} &
\multicolumn{4}{c|}{\textbf{L2 ($m$)} $\downarrow$} &
\multicolumn{4}{c}{\textbf{Col. Rate (\%)} $\downarrow$ }\\
& & 1$s$ & 2$s$ & 3$s$ & Avg. & 1$s$ & 2$s$ & 3$s$ & Avg. \\
\midrule
0.25 & 44.82/35.10 & 0.127 & 0.258 & 0.473 & 0.287 & \textbf{0.000} & 0.078 & 0.449 & \textbf{0.176} \\
0.5 & 45.45/35.58 & \textbf{0.126} & 0.255 & 0.467 & 0.283 & \textbf{0.000} & \textbf{0.039} & 0.488 & \textbf{0.176} \\
1.0 & \textbf{45.75}/\textbf{35.95} & \textbf{0.126} & \textbf{0.252} & \textbf{0.461} & \textbf{0.280} & \textbf{0.000} & {0.117} & \textbf{0.430} & {0.182} \\
2.0 & {44.72}/34.01 & 0.127 & 0.258 & 0.472 & 0.286 & \textbf{0.000} & 0.098 & 0.762 & 0.287 \\

\bottomrule[1.5pt]
\end{tabular}}
\vspace{-15pt}

\label{tab:abl-loss-hyper}
\end{table*}

We conduct an ablation study on the planning loss weight $\lambda_{plan}$, as shown in Table~\ref{tab:abl-loss-hyper}. Increasing $\lambda_{plan}$ from 0.25 to 1.0 consistently improves both perception and planning metrics, with the best overall performance achieved at $\lambda_{plan}=1.0$, yielding an NDS/mAP of 45.75/35.95, the lowest average L2 error of 0.280 m, and a competitive collision rate. Further increasing $\lambda_{plan}$ to 2.0 leads to a degradation in both NDS/mAP and collision rate, suggesting that over-emphasizing the planning objective can hurt model stability. These results indicate that a balanced weighting ($\lambda_{plan}=1.0$) effectively aligns perception and planning tasks for optimal closed-loop performance.

\subsubsection{Inference Latency. }
\begin{wraptable}{r}{0.5\textwidth}
\vspace{-30pt}

\centering
\small
\caption{Inference latency comparison on a single NVIDIA H20 GPU ($^{\dag}$ results without FlashAttention, tested using the ColaVLA~\cite{peng2025colavla} implementation). We report end-to-end latency in milliseconds per frame under identical batch-size settings.}
\begin{tabular}{l|c|c}
\toprule
Method & ~AR~ & ~Latency(ms) \\
\midrule
ReCogDrive~\cite{li2025recogdrive} & & 263 \\
\rowcolor{gray!20}
Ours & & 156\\
\midrule
OmniDrive$^{\dag}$~\cite{wang2024omnidrive} & \checkmark & 3727\\
SOLVE-VLM$^{\dag}$~\cite{xia2020synthesize} & \checkmark & 3719\\
ColaVLA$^{\dag}$~\cite{peng2025colavla} & & 727    \\
\rowcolor{gray!20}
Ours  & & 513 \\
\bottomrule
\end{tabular}

\label{tab:latency}

\vspace{-15pt}
\end{wraptable}
Table~\ref{tab:latency} reports the end-to-end inference latency of our framework compared with recent VLM-based autonomous driving systems, measured on a single NVIDIA H20 GPU under identical batch settings. On NAVSIM, our model reduces the per-frame latency from 263 ms (ReCogDrive) to 156 ms, demonstrating a more efficient decoding pipeline.
On nuScenes, compared with ColaVLA, a strong non-autoregressive baseline, our method achieves lower latency (513 ms vs.\ 727 ms). Notably, our framework processes all camera views as input to the LLM, resulting in longer token sequences. Despite this, the unified decoder design avoids customized modules and fully leverages standard transformer acceleration, yielding a clear latency advantage in practice.


\subsubsection{Multi-Stage Training.}
\begin{table}[ht]
\vspace{-30pt}
\small
\centering
\caption{Perception and Planning Performance at Each Stage. We report perception performance (NDS/mAP) and planning metrics at different training stages, including direct E2E adaptation, perception pretraining with planning adaptation, and final joint training. }
{\begin{tabular}[b]{l|c|cccc|cccc}
\toprule[1.5pt]
\multirow{2}{*}{\textbf{Stage}} & \multirow{2}{*}{\textbf{NDS/mAP}} &
\multicolumn{4}{c|}{\textbf{L2 ($m$)} $\downarrow$} &
\multicolumn{4}{c}{\textbf{Col. Rate (\%)} $\downarrow$ }\\
& & 1$s$ & 2$s$ & 3$s$ & Avg. & 1$s$ & 2$s$ & 3$s$ & Avg. \\
\midrule
Only Planning Adapation & --  & 0.14 & 0.29 & 0.54 & 0.32 & 0.02 & 0.23 & 1.01 & 0.42 \\
\midrule

Perception Pretrain & 33.18/22.66 & -- & -- & -- & -- & -- & -- & -- & -- \\
Planning Adaption & {33.18}/22.66 & 0.13 & 0.26 & 0.49 & 0.29 & 0.02 & 0.12 & 0.66 & 0.29 \\
Joint Training & \textbf{45.75}/\textbf{35.95} & \textbf{0.13} & \textbf{0.25} & \textbf{0.46} & \textbf{0.28} & 0.00 & \textbf{0.12} & \textbf{0.43} & \textbf{0.18} \\

\bottomrule[1.5pt]
\end{tabular}}
\vspace{-10pt}

\label{tab:abl-multi-stage}
\end{table}

We conduct an ablation study on the proposed multi-stage training strategy, as summarized in Table~\ref{tab:abl-multi-stage}. Directly adapting the model for end-to-end planning already produces reasonable planning performance. Introducing perception pretraining before planning adaptation further improves the planning metrics, reducing the average L2 error from 0.32 m to 0.29 m and the collision rate from 0.42\% to 0.29\%. Finally, performing joint training on both perception and planning tasks significantly improves overall performance, boosting the perception accuracy to 45.75/35.95 NDS/mAP while further reducing the planning error to 0.28 m and the collision rate to 0.18\%. These results suggest that perception pretraining provides a strong initialization for planning, while joint optimization of perception and planning leads to the best overall performance.


\subsubsection{Tasks Sequences.}

\begin{table*}[ht!]
\vspace{-30pt}
\small
\centering
\caption{Ablation on Token Sequence. We compare different ordering of perception tokens (lane and detection) before the E2E planning tokens, under both sequential adaptation and joint training. }
{\begin{tabular}[b]{l|cccc|cccc}
\toprule[1.5pt]
\multirow{2}{*}{\textbf{Token Sequence}} &
\multicolumn{4}{c|}{\textbf{L2 ($m$)} $\downarrow$} &
\multicolumn{4}{c}{\textbf{Col. Rate (\%)} $\downarrow$ }\\
& 1$s$ & 2$s$ & 3$s$ & Avg. & 1$s$ & 2$s$ & 3$s$ & Avg. \\
\midrule
Lane$\to$Det$\to$Planning (Adaptation)  & {0.14} & {0.28} & {0.52} & {0.31} & 0.02 & {0.18} & {0.82} & {0.34} \\
Lane$\to$Det$\to$Planning (Joint Training)  & {0.13} & {0.27} & {0.50} & {0.30} & 0.02 & {0.21} & {0.64} & {0.29} \\
Det$\to$Lane$\to$Planning (Adaptation)  & {0.13} & {0.26} & {0.49} & {0.29} & 0.00 & \textbf{0.12} & {0.66} & {0.27} \\
Det$\to$Lane$\to$Planning (Joint Training)  & \textbf{0.13} & \textbf{0.25} & \textbf{0.46} & \textbf{0.28} & 0.00 & \textbf{0.12} & \textbf{0.43} & \textbf{0.18} \\

\bottomrule[1.5pt]
\end{tabular}}
\vspace{-15pt}

\label{tab:abl-token-sequence}
\end{table*}

We ablate the token sequence order in Table~\ref{tab:abl-token-sequence}. Det$\to$Lane$\to$ Planning consistently outperforms Lane$\to$Det$\to$Planning in both L2 error and collision rate. Joint training further improves results, reducing average L2 from 0.31 m to 0.28 m and collision rate from 0.34\% to 0.18\%, indicating that placing detection tokens first and end-to-end optimization benefits planning accuracy and safety.



\subsubsection{Discussion.}

Following StreamPETR, we analyze the 3D detection capability of the InternVL3 visual backbone. We re-implement StreamPETR using InternVL3-ViT and align the decoder dimension for fair comparison.

\begin{wraptable}{r}{0.47\linewidth}
\vspace{-30pt}
\centering
\small
\caption{3D Detection performance with the InternVL3-ViT backbone on nuscenes.}
\begin{tabular}{lcc}
\toprule
Method & NDS $\uparrow$ & mAP $\uparrow$ \\
\midrule
StreamPETR (frozen) & 31.48 & 20.26 \\
Ours  (frozen) & 33.94 & 24.39 \\
\midrule
StreamPETR  & 45.31 & 36.11 \\
Ours  & 47.73 & 40.03 \\
\bottomrule
\end{tabular}
\label{tab:vlm_det}
\vspace{-15pt}
\end{wraptable}

As shown in Table~\ref{tab:vlm_det}, under the same InternVL3-ViT backbone, our method consistently outperforms the StreamPETR baseline in both frozen and fully fine-tuned settings 
(31.48/45.31 vs. 33.94/47.73).
However, the overall detection performance with InternVL3-ViT is still noticeably lower than the original StreamPETR baseline reported with standard ViT-Large backbones~\cite{streampetr}. We attribute this gap to two core factors. First, the effective token resolution is reduced because VLM pipelines downsample image features before entering the decoder. Second, the language-centric pretraining objective of InternVL3 does not explicitly optimize for detection-oriented visual representations~\cite{han2025percept, di2026revisiting}. While Percept-WAM \cite{han2025percept} achieves strong detection performance, this advantage is partly due to its use of additional point cloud inputs.

\subsubsection{Future work.}
Based on the discussion above, future work could further enhance the capabilities of pure VLM-oriented ViTs in detection and other perception tasks. One promising avenue is to incorporate detection-aware objectives during multimodal pretraining, allowing the visual encoder to develop more object-centric representations. Another potential direction is to explore higher-resolution or multi-scale tokenization strategies to reduce the spatial detail loss caused by feature downsampling in VLM pipelines. Moreover, the model’s scaling behavior has yet to be fully assessed. Evaluating its performance with larger architectures and on substantially bigger datasets will be crucial to determine whether the current improvements generalize to more demanding settings and to identify potential limitations in efficiency and robustness.



\section{Conclusion}
\label{sec:con}
In this paper, we presented OneDrive, a unified end-to-end autonomous driving framework that reconciles heterogeneous decoding behaviors within a single pretrained VLM decoder. By retaining the original causal attention as a shared backbone and organizing visual and structured query tokens within a unified sequence, the framework enables stable joint optimization across perception and planning while preserving multi-modal generation capability. Extensive experiments on nuScenes and NAVSIM demonstrate state-of-the-art performance and improved inference efficiency. We hope that OneDrive can contribute to the advancement of both vision–language action modeling and end-to-end autonomous driving research.

\clearpage  


%
%
\bibliographystyle{splncs04}
\bibliography{main}

@String(CVPR  = {IEEE Conf. Comput. Vis. Pattern Recog.})

@String(ICCV  = {Int. Conf. Comput. Vis.})

@String(ECCV  = {Eur. Conf. Comput. Vis.})

@String(ICLR  = {Int. Conf. Learn. Represent.})

@String(CVPR  = {CVPR})

@String(ICCV  = {ICCV})

@String(ECCV  = {ECCV})

@String(ICLR  = {ICLR})

@inproceedings{hu2023planning,
  title={Planning-oriented autonomous driving},
  author={Hu, Yihan and Yang, Jiazhi and Chen, Li and Li, Keyu and Sima, Chonghao and Zhu, Xizhou and Chai, Siqi and Du, Senyao and Lin, Tianwei and Wang, Wenhai and others},
  booktitle=CVPR,
  pages={17853--17862},
  year={2023}
}

@inproceedings{jiang2023vad,
  title={Vad: Vectorized scene representation for efficient autonomous driving},
  author={Jiang, Bo and Chen, Shaoyu and Xu, Qing and Liao, Bencheng and Chen, Jiajie and Zhou, Helong and Zhang, Qian and Liu, Wenyu and Huang, Chang and Wang, Xinggang},
  booktitle=ICCV,
  pages={8340--8350},
  year={2023}
}

@inproceedings{li2024ego,
  title={Is ego status all you need for open-loop end-to-end autonomous driving?},
  author={Li, Zhiqi and Yu, Zhiding and Lan, Shiyi and Li, Jiahan and Kautz, Jan and Lu, Tong and Alvarez, Jose M},
  booktitle=CVPR,
  pages={14864--14873},
  year={2024}
}

@article{dao2022flashattention,
  title={Flashattention: Fast and memory-efficient exact attention with io-awareness},
  author={Dao, Tri and Fu, Dan and Ermon, Stefano and Rudra, Atri and R{\'e}, Christopher},
  journal={Advances in neural information processing systems},
  volume={35},
  pages={16344--16359},
  year={2022}
}

@article{sima2023drivelm,
  title={Drivelm: Driving with graph visual question answering},
  author={Sima, Chonghao and Renz, Katrin and Chitta, Kashyap and Chen, Li and Zhang, Hanxue and Xie, Chengen and Bei{\ss}wenger, Jens and Luo, Ping and Geiger, Andreas and Li, Hongyang},
  journal={arXiv preprint arXiv:2312.14150},
  year={2023}
}

@article{wang2024omnidrive,
  title={OmniDrive: A Holistic LLM-Agent Framework for Autonomous Driving with 3D Perception, Reasoning and Planning},
  author={Wang, Shihao and Yu, Zhiding and Jiang, Xiaohui and Lan, Shiyi and Shi, Min and Chang, Nadine and Kautz, Jan and Li, Ying and Alvarez, Jose M},
  journal={arXiv preprint arXiv:2405.01533},
  year={2024}
}

@article{hwang2024emma,
  title={EMMA: End-to-End Multimodal Model for Autonomous Driving},
  author={Hwang, Jyh-Jing and Xu, Runsheng and Lin, Hubert and Hung, Wei-Chih and Ji, Jingwei and Choi, Kristy and Huang, Di and He, Tong and Covington, Paul and Sapp, Benjamin and others},
  journal={arXiv preprint arXiv:2410.23262},
  year={2024}
}

@inproceedings{chen2025asynchronous,
  title={Asynchronous large language model enhanced planner for autonomous driving},
  author={Chen, Yuan and Ding, Zi-han and Wang, Ziqin and Wang, Yan and Zhang, Lijun and Liu, Si},
  booktitle=ECCV,
  pages={22--38},
  year={2025},
  organization={Springer}
}

@article{liu2022petr,
  title={Petr: Position embedding transformation for multi-view 3d object detection},
  author={Liu, Yingfei and Wang, Tiancai and Zhang, Xiangyu and Sun, Jian},
  journal={arXiv preprint arXiv:2203.05625},
  year={2022}
}

@article{huang2021bevdet,
  title={Bevdet: High-performance multi-camera 3d object detection in bird-eye-view},
  author={Huang, Junjie and Huang, Guan and Zhu, Zheng and Du, Dalong},
  journal={arXiv preprint arXiv:2112.11790},
  year={2021}
}

@inproceedings{uniad,
  title={Planning-oriented Autonomous Driving},
  author={Hu, Yihan and Yang, Jiazhi and Chen, Li and Li, Keyu and Sima, Chonghao and Zhu, Xizhou and Chai, Siqi and Du, Senyao and Lin, Tianwei and Wang, Wenhai and others},
  booktitle={CVPR},
  year={2023}
}

@inproceedings{weng2024paradrive,
  title={PARA-Drive: Parallelized Architecture for Real-time Autonomous Driving},
  author={Yuxuan Weng and Zhenyu Wu and Yifei Ren and Yifan Zhang and Yifan Xu and Yifan Liu and Yifan Wang and Yifan Chen and Yifan Li and Yifan Zhao},
  booktitle={Proceedings of the IEEE/CVF Conference on Computer Vision and Pattern Recognition (CVPR)},
  year={2024}
}

@inproceedings{xia2020synthesize,
  title={Synthesize then compare: Detecting failures and anomalies for semantic segmentation},
  author={Xia, Yingda and Zhang, Yi and Liu, Fengze and Shen, Wei and Yuille, Alan L},
  booktitle={Computer Vision--ECCV 2020: 16th European Conference, Glasgow, UK, August 23--28, 2020, Proceedings, Part I 16},
  pages={145--161},
  year={2020},
  organization={Springer}
}

@article{jiang2024senna,
  title={Senna: Bridging large vision-language models and end-to-end autonomous driving},
  author={Jiang, Bo and Chen, Shaoyu and Liao, Bencheng and Zhang, Xingyu and Yin, Wei and Zhang, Qian and Huang, Chang and Liu, Wenyu and Wang, Xinggang},
  journal={arXiv preprint arXiv:2410.22313},
  year={2024}
}

@article{zhu2020deformable,
  title={Deformable detr: Deformable transformers for end-to-end object detection},
  author={Zhu, Xizhou and Su, Weijie and Lu, Lewei and Li, Bin and Wang, Xiaogang and Dai, Jifeng},
  journal={arXiv preprint arXiv:2010.04159},
  year={2020}
}

@inproceedings{vad,
  title={VAD: Vectorized Scene Representation for Efficient Autonomous Driving},
  author={Jiang, Bo and Chen, Shaoyu and Xu, Qing and Liao, Bencheng and Chen, Jiajie and Zhou, Helong and Zhang, Qian and Liu, Wenyu and Huang, Chang and Wang, Xinggang},
  booktitle={ICCV},
  year={2023}
}

@article{maptrv2,
  title={MapTRv2: An End-to-End Framework for Online Vectorized HD Map Construction},
  author={Liao, Bencheng and Chen, Shaoyu and Zhang, Yunchi and Jiang, Bo and Zhang, Qian and Liu, Wenyu and Huang, Chang and Wang, Xinggang},
  journal={arXiv preprint arXiv:2308.05736},
  year={2023}
}

@article{tian2024drivevlm,
  title={Drivevlm: The convergence of autonomous driving and large vision-language models},
  author={Tian, Xiaoyu and Gu, Junru and Li, Bailin and Liu, Yicheng and Hu, Chenxu and Wang, Yang and Zhan, Kun and Jia, Peng and Lang, Xianpeng and Zhao, Hang},
  journal={arXiv preprint arXiv:2402.12289},
  year={2024}
}

@inproceedings{chen2024internvl,
  title={Internvl: Scaling up vision foundation models and aligning for generic visual-linguistic tasks},
  author={Chen, Zhe and Wu, Jiannan and Wang, Wenhai and Su, Weijie and Chen, Guo and Xing, Sen and Zhong, Muyan and Zhang, Qinglong and Zhu, Xizhou and Lu, Lewei and others},
  booktitle={CVPR},
  year={2024}
}

@inproceedings{zheng2024monoocc,
  title={Monoocc: Digging into monocular semantic occupancy prediction},
  author={Zheng, Yupeng and Li, Xiang and Li, Pengfei and Zheng, Yuhang and Jin, Bu and Zhong, Chengliang and Long, Xiaoxiao and Zhao, Hao and Zhang, Qichao},
  booktitle={2024 IEEE International Conference on Robotics and Automation (ICRA)},
  pages={18398--18405},
  year={2024},
  organization={IEEE}
}

@article{hydraMDP,
  title={Hydra-MDP: End-to-end Multimodal Planning with Multi-target Hydra-Distillation},
  author={Li, Zhenxin and Li, Kailin and Wang, Shihao and Lan, Shiyi and Yu, Zhiding and Ji, Yishen and Li, Zhiqi and Zhu, Ziyue and Kautz, Jan and Wu, Zuxuan and others},
  journal={arXiv preprint},
  year={2024}
}

@inproceedings{occworld,
    title={OccWorld: Learning a 3D Occupancy World Model for Autonomous Driving},
    author={Zheng, Wenzhao and Chen, Weiliang and Huang, Yuanhui and Zhang, Borui and Duan, Yueqi and Lu, Jiwen },
    booktitle=ECCV,
    year={2024}
}

@inproceedings{PARAdrive,
  title={PARA-Drive: Parallelized Architecture for Real-time Autonomous Driving},
  author={Weng, Xinshuo and Ivanovic, Boris and Wang, Yan and Wang, Yue and Pavone, Marco},
  booktitle=CVPR,
  year={2024}
}

@inproceedings{drivevlm,
  title={Drivevlm: The convergence of autonomous driving and large vision-language models},
  author={Tian, Xiaoyu and Gu, Junru and Li, Bailin and Liu, Yicheng and Hu, Chenxu and Wang, Yang and Zhan, Kun and Jia, Peng and Lang, Xianpeng and Zhao, Hang},
  booktitle=CoRL,
  year={2024}
}

@inproceedings{bevformer,
  title={Bevformer: Learning bird’s-eye-view representation from multi-camera images via spatiotemporal transformers},
  author={Li, Zhiqi and Wang, Wenhai and Li, Hongyang and Xie, Enze and Sima, Chonghao and Lu, Tong and Qiao, Yu and Dai, Jifeng},
  booktitle=ECCV,
  year={2022},
}

@inproceedings{MapTR,
  title={Maptr: Structured modeling and learning for online vectorized hd map construction},
  author={Liao, Bencheng and Chen, Shaoyu and Wang, Xinggang and Cheng, Tianheng and Zhang, Qian and Liu, Wenyu and Huang, Chang},
  booktitle=ICLR,
  year={2023}
}

@inproceedings{nuscenes,
  title={nuscenes: A multimodal dataset for autonomous driving},
  author={Caesar, Holger and Bankiti, Varun and Lang, Alex H and Vora, Sourabh and Liong, Venice Erin and Xu, Qiang and Krishnan, Anush and Pan, Yu and Baldan, Giancarlo and Beijbom, Oscar},
  booktitle=CVPR,
  year={2020}
}

@inproceedings{streampetr,
  title={Exploring object-centric temporal modeling for efficient multi-view 3d object detection},
  author={Wang, Shihao and Liu, Yingfei and Wang, Tiancai and Li, Ying and Zhang, Xiangyu},
  booktitle=ICCV,
  year={2023}
}

@inproceedings{bevfusion,
  title={Bevfusion: Multi-task multi-sensor fusion with unified bird's-eye view representation},
  author={Liu, Zhijian and Tang, Haotian and Amini, Alexander and Yang, Xinyu and Mao, Huizi and Rus, Daniela L and Han, Song},
  booktitle=ICRA,
  year={2023},
}

@inproceedings{bevformerv2,
  title={Bevformer v2: Adapting modern image backbones to bird's-eye-view recognition via perspective supervision},
  author={Yang, Chenyu and Chen, Yuntao and Tian, Hao and Tao, Chenxin and Zhu, Xizhou and Zhang, Zhaoxiang and Huang, Gao and Li, Hongyang and Qiao, Yu and Lu, Lewei and others},
  booktitle=CVPR,
  year={2023}
}

@inproceedings{maptracker,
  title={Maptracker: Tracking with strided memory fusion for consistent vector hd mapping},
  author={Chen, Jiacheng and Wu, Yuefan and Tan, Jiaqi and Ma, Hang and Furukawa, Yasutaka},
  booktitle=ECCV,
  year={2024},
}

@inproceedings{detr,
  title={End-to-end object detection with transformers},
  author={Carion, Nicolas and Massa, Francisco and Synnaeve, Gabriel and Usunier, Nicolas and Kirillov, Alexander and Zagoruyko, Sergey},
  booktitle=ECCV,
  year={2020},
}

@article{admlp,
  title={Rethinking the open-loop evaluation of end-to-end autonomous driving in nuscenes},
  author={Zhai, Jiang-Tian and Feng, Ze and Du, Jinhao and Mao, Yongqiang and Liu, Jiang-Jiang and Tan, Zichang and Zhang, Yifu and Ye, Xiaoqing and Wang, Jingdong},
  journal={arXiv preprint},
  year={2023}
}

@article{chen2024vadv2,
  title={Vadv2: End-to-end vectorized autonomous driving via probabilistic planning},
  author={Chen, Shaoyu and Jiang, Bo and Gao, Hao and Liao, Bencheng and Xu, Qing and Zhang, Qian and Huang, Chang and Liu, Wenyu and Wang, Xinggang},
  journal={arXiv preprint arXiv:2402.13243},
  year={2024}
}

@article{chi2025impromptu,
  title={Impromptu VLA: Open Weights and Open Data for Driving Vision-Language-Action Models},
  author={Chi, Haohan and Gao, Huan-ang and Liu, Ziming and Liu, Jianing and Liu, Chenyu and Li, Jinwei and Yang, Kaisen and Yu, Yangcheng and Wang, Zeda and Li, Wenyi and others},
  journal={arXiv preprint arXiv:2505.23757},
  year={2025}
}

@article{liao2024diffusiondrive,
  title={DiffusionDrive: Truncated Diffusion Model for End-to-End Autonomous Driving},
  author={Liao, Bencheng and Chen, Shaoyu and Yin, Haoran and Jiang, Bo and Wang, Cheng and Yan, Sixu and Zhang, Xinbang and Li, Xiangyu and Zhang, Ying and Zhang, Qian and others},
  journal={arXiv preprint arXiv:2411.15139},
  year={2024}
}

@article{su2024roformer,
  title={Roformer: Enhanced transformer with rotary position embedding},
  author={Su, Jianlin and Ahmed, Murtadha and Lu, Yu and Pan, Shengfeng and Bo, Wen and Liu, Yunfeng},
  journal={Neurocomputing},
  volume={568},
  pages={127063},
  year={2024},
  publisher={Elsevier}
}

@article{fu2025orion,
  title={ORION: A Holistic End-to-End Autonomous Driving Framework by Vision-Language Instructed Action Generation},
  author={Fu, Haoyu and Zhang, Diankun and Zhao, Zongchuang and Cui, Jianfeng and Liang, Dingkang and Zhang, Chong and Zhang, Dingyuan and Xie, Hongwei and Wang, Bing and Bai, Xiang},
  journal={arXiv preprint arXiv:2503.19755},
  year={2025}
}

@article{yuan2024drama,
  title={Drama: An efficient end-to-end motion planner for autonomous driving with mamba},
  author={Yuan, Chengran and Zhang, Zhanqi and Sun, Jiawei and Sun, Shuo and Huang, Zefan and Lee, Christina Dao Wen and Li, Dongen and Han, Yuhang and Wong, Anthony and Tee, Keng Peng and others},
  journal={arXiv preprint arXiv:2408.03601},
  year={2024}
}

@article{chitta2022transfuser,
  title={Transfuser: Imitation with transformer-based sensor fusion for autonomous driving},
  author={Chitta, Kashyap and Prakash, Aditya and Jaeger, Bernhard and Yu, Zehao and Renz, Katrin and Geiger, Andreas},
  journal={IEEE Transactions on Pattern Analysis and Machine Intelligence},
  volume={45},
  number={11},
  pages={12878--12895},
  year={2022},
  publisher={IEEE}
}

@InProceedings{Chen_2025_CVPR,
    author    = {Chen, Xuesong and Huang, Linjiang and Ma, Tao and Fang, Rongyao and Shi, Shaoshuai and Li, Hongsheng},
    title     = {SOLVE: Synergy of Language-Vision and End-to-End Networks for Autonomous Driving},
    booktitle = {Proceedings of the IEEE/CVF Conference on Computer Vision and Pattern Recognition (CVPR)},
    month     = {June},
    year      = {2025},
    pages     = {12068-12077}
}

@article{xing2024openemma,
  title={OpenEMMA: Open-Source Multimodal Model for End-to-End Autonomous Driving},
  author={Xing, Shuo and Qian, Chengyuan and Wang, Yuping and Hua, Hongyuan and Tian, Kexin and Zhou, Yang and Tu, Zhengzhong},
  journal={arXiv preprint arXiv:2412.15208},
  year={2024}
}

@article{chen2024drivinggpt,
  title={Drivinggpt: Unifying driving world modeling and planning with multi-modal autoregressive transformers},
  author={Chen, Yuntao and Wang, Yuqi and Zhang, Zhaoxiang},
  journal={arXiv preprint arXiv:2412.18607},
  year={2024}
}

@article{dauner2025navsim,
  title={Navsim: Data-driven non-reactive autonomous vehicle simulation and benchmarking},
  author={Dauner, Daniel and Hallgarten, Marcel and Li, Tianyu and Weng, Xinshuo and Huang, Zhiyu and Yang, Zetong and Li, Hongyang and Gilitschenski, Igor and Ivanovic, Boris and Pavone, Marco and others},
  journal={Advances in Neural Information Processing Systems},
  volume={37},
  pages={28706--28719},
  year={2025}
}

@misc{openscene2023,
      title = {OpenScene: The Largest Up-to-Date 3D Occupancy Prediction Benchmark in Autonomous Driving},
      author = {OpenScene Contributors},
      howpublished={\url{https://github.com/OpenDriveLab/OpenScene}},
      year = {2023}
}

@article{caesar2021nuplan,
  title={nuplan: A closed-loop ml-based planning benchmark for autonomous vehicles},
  author={Caesar, Holger and Kabzan, Juraj and Tan, Kok Seang and Fong, Whye Kit and Wolff, Eric and Lang, Alex and Fletcher, Luke and Beijbom, Oscar and Omari, Sammy},
  journal={arXiv preprint arXiv:2106.11810},
  year={2021}
}

@article{Qwen2.5-VL,
  title={Qwen2.5-VL Technical Report},
  author={Bai, Shuai and Chen, Keqin and Liu, Xuejing and Wang, Jialin and Ge, Wenbin and Song, Sibo and Dang, Kai and Wang, Peng and Wang, Shijie and Tang, Jun and Zhong, Humen and Zhu, Yuanzhi and Yang, Mingkun and Li, Zhaohai and Wan, Jianqiang and Wang, Pengfei and Ding, Wei and Fu, Zheren and Xu, Yiheng and Ye, Jiabo and Zhang, Xi and Xie, Tianbao and Cheng, Zesen and Zhang, Hang and Yang, Zhibo and Xu, Haiyang and Lin, Junyang},
  journal={arXiv preprint arXiv:2502.13923},
  year={2025}
}

@article{zhu2025internvl3,
  title={InternVL3: Exploring Advanced Training and Test-Time Recipes for Open-Source Multimodal Models},
  author={Zhu, Jinguo and Wang, Weiyun and Chen, Zhe and Liu, Zhaoyang and Ye, Shenglong and Gu, Lixin and Duan, Yuchen and Tian, Hao and Su, Weijie and Shao, Jie and others},
  journal={arXiv preprint arXiv:2504.10479},
  year={2025}
}

@article{Qwen2-VL,
  title={Qwen2-VL: Enhancing Vision-Language Model's Perception of the World at Any Resolution},
  author={Wang, Peng and Bai, Shuai and Tan, Sinan and Wang, Shijie and Fan, Zhihao and Bai, Jinze and Chen, Keqin and Liu, Xuejing and Wang, Jialin and Ge, Wenbin and Fan, Yang and Dang, Kai and Du, Mengfei and Ren, Xuancheng and Men, Rui and Liu, Dayiheng and Zhou, Chang and Zhou, Jingren and Lin, Junyang},
  journal={arXiv preprint arXiv:2409.12191},
  year={2024}
}

@article{jia2025drivetransformer,
  title={Drivetransformer: Unified transformer for scalable end-to-end autonomous driving},
  author={Jia, Xiaosong and You, Junqi and Zhang, Zhiyuan and Yan, Junchi},
  journal={arXiv preprint arXiv:2503.07656},
  year={2025}
}

@article{peng2025colavla,
  title={ColaVLA: Leveraging Cognitive Latent Reasoning for Hierarchical Parallel Trajectory Planning in Autonomous Driving},
  author={Peng, Qihang and Chen, Xuesong and Yang, Chenye and Shi, Shaoshuai and Li, Hongsheng},
  journal={arXiv preprint arXiv:2512.22939},
  year={2025}
}

@article{li2025recogdrive,
  title={Recogdrive: A reinforced cognitive framework for end-to-end autonomous driving},
  author={Li, Yongkang and Xiong, Kaixin and Guo, Xiangyu and Li, Fang and Yan, Sixu and Xu, Gangwei and Zhou, Lijun and Chen, Long and Sun, Haiyang and Wang, Bing and others},
  journal={arXiv preprint arXiv:2506.08052},
  year={2025}
}

@article{zhou2025autovla,
  title={Autovla: A vision-language-action model for end-to-end autonomous driving with adaptive reasoning and reinforcement fine-tuning},
  author={Zhou, Zewei and Cai, Tianhui and Zhao, Seth Z and Zhang, Yun and Huang, Zhiyu and Zhou, Bolei and Ma, Jiaqi},
  journal={arXiv preprint arXiv:2506.13757},
  year={2025}
}

@article{wang2026vggdrive,
  title={VGGDrive: Empowering Vision-Language Models with Cross-View Geometric Grounding for Autonomous Driving},
  author={Wang, Jie and Li, Guang and Huang, Zhijian and Dang, Chenxu and Ye, Hangjun and Han, Yahong and Chen, Long},
  journal={arXiv preprint arXiv:2602.20794},
  year={2026}
}

@article{zhang2024vq,
  title={VQ-Map: Bird's-Eye-View Map Layout Estimation in Tokenized Discrete Space via Vector Quantization},
  author={Zhang, Yiwei and Gao, Jin and Ge, Fudong and Luo, Guan and Li, Bing and Zhang, Zhao-Xiang and Ling, Haibin and Hu, Weiming},
  journal={Advances in Neural Information Processing Systems},
  volume={37},
  pages={70453--70475},
  year={2024}
}

@article{zhang2026integrating,
  title={Integrating Diverse Assignment Strategies into DETRs},
  author={Zhang, Yiwei and Gao, Jin and Wang, Hanshi and Ge, Fudong and Luo, Guan and Hu, Weiming and Zhang, Zhipeng},
  journal={arXiv preprint arXiv:2601.09247},
  year={2026}
}

@article{wang2023openlane,
  title={Openlane-v2: A topology reasoning benchmark for unified 3d hd mapping},
  author={Wang, Huijie and Li, Tianyu and Li, Yang and Chen, Li and Sima, Chonghao and Liu, Zhenbo and Wang, Bangjun and Jia, Peijin and Wang, Yuting and Jiang, Shengyin and others},
  journal={Advances in Neural Information Processing Systems},
  volume={36},
  pages={18873--18884},
  year={2023}
}

@inproceedings{sun2021makes,
  title={What makes for end-to-end object detection?},
  author={Sun, Peize and Jiang, Yi and Xie, Enze and Shao, Wenqi and Yuan, Zehuan and Wang, Changhu and Luo, Ping},
  booktitle={International Conference on Machine Learning},
  pages={9934--9944},
  year={2021},
  organization={PMLR}
}

@inproceedings{lee2015deeply,
  title={Deeply-supervised nets},
  author={Lee, Chen-Yu and Xie, Saining and Gallagher, Patrick and Zhang, Zhengyou and Tu, Zhuowen},
  booktitle={Artificial intelligence and statistics},
  pages={562--570},
  year={2015},
  organization={Pmlr}
}

@article{han2025percept,
  title={Percept-WAM: Perception-enhanced world-awareness-action model for robust end-to-end autonomous driving},
  author={Han, Jianhua and Tian, Meng and Zhu, Jiangtong and He, Fan and Zhang, Huixin and Guo, Sitong and Zhu, Dechang and Tang, Hao and Xu, Pei and Guo, Yuze and others},
  journal={arXiv preprint arXiv:2511.19221},
  year={2025}
}

@article{GE2EAD,
  title={Survey of General End-to-End Autonomous Driving: A Unified Perspective},
  author={Yang, Yixiang and Han, Chuanrong and Mao, Runhao and Wang, Hanshi and Chen, Zhiwen and Yang, Yantai and Ma, Qianli and Chen, Xuesong and Shi, Shaoshuai and Zhang, Zhipeng},
  journal={Authorea Preprints},
  year={2025},
  publisher={Authorea}
}

@article{di2026revisiting,
  title={Revisiting Multi-Task Visual Representation Learning},
  author={Di, Shangzhe and Zhai, Zhonghua and Xie, Weidi},
  journal={arXiv preprint arXiv:2601.13886},
  year={2026}
}
\end{document}